\documentclass{SCIS2023}
\usepackage[table,xcdraw]{xcolor}
\usepackage{enumitem}
\usepackage{multirow}
\usepackage{graphicx}
\usepackage{wrapfig}
\usepackage[version=4]{mhchem}

\begin{document}

\ArticleType{RESEARCH PAPER}
\Year{2022}
\Month{}
\Vol{}
\No{}
\DOI{}
\ArtNo{}
\ReceiveDate{}
\ReviseDate{}
\AcceptDate{}
\OnlineDate{}

\title{Careful Selection and Thoughtful Discarding: Graph Explicit Pooling Utilizing Discarded Nodes}

\author[1]{Chuang LIU}{}
\author[1]{Wenhang YU}{}
\author[1]{Kuang GAO}{}
\author[2]{Xueqi MA}{}
\author[3]{Yibing ZHAN}{}
\author[4]{\\Jia WU}{}
\author[1]{Bo DU}{}
\author[1]{Wenbin HU}{{hwb@whu.edu.cn}}




\address[1]{School of Computer Science, Wuhan University, Wuhan {\rm 430072}, China}
\address[2]{School of Computing and Information Systems, The University of Melbourne,  Melbourne {\rm 3052}, Australia}
\address[3]{JD Explore Academy, Beijing {\rm 100176}, China}
\address[4]{School of Computing, Macquarie University,  Sydney {\rm 2109}, Australia}

\abstract{Graph pooling has been increasingly recognized as crucial for Graph Neural Networks (GNNs) to facilitate hierarchical graph representation learning. Existing graph pooling methods commonly consist of two stages: selecting top-ranked nodes and discarding the remaining to construct coarsened graph representations. However, this paper highlights two key issues with these methods: \textbf{1)} The process of selecting nodes to discard frequently employs additional Graph Convolutional Networks or Multilayer Perceptrons, lacking a thorough evaluation of each node's impact on the final graph representation and subsequent prediction tasks. \textbf{2)} Current graph pooling methods tend to directly discard the noise segment (dropped) of the graph without accounting for the latent information contained within these elements. To address the \textbf{first} issue, we introduce a novel \underline{Gr}aph \underline{e}xplicit \underline{Pool}ing (GrePool) method, which selects nodes by explicitly leveraging the relationships between the nodes and final representation vectors crucial for classification. The \textbf{second} issue is addressed using an extended version of GrePool (\textit{i.e.}, GrePool+), which applies a uniform loss on the discarded nodes. This addition is designed  to augment the training process and improve classification accuracy.  Furthermore, we conduct comprehensive experiments across 12 widely used datasets to validate our proposed method’s effectiveness, including the Open Graph Benchmark datasets. Our experimental results uniformly demonstrate that GrePool outperforms 14 baseline methods for most datasets. Likewise, implementing GrePool+ enhances GrePool's performance without incurring additional computational costs.

}


\maketitle

\section{Introduction}

Graph Neural Networks (GNNs) ~\cite{gcn,scis-deepwalk,scis-grarf} have consistently displayed remarkable performance for several graph classification tasks, including predicting molecular properties, diagnosing cancer, and analyzing brain data~\cite{fair-graph-classification, gmt}. In comparison to node-level tasks, such as node classification, which predominantly utilize Graph Convolutional Networks (GCNs)~\cite{gcn} to create node representations for subsequent tasks, graph classification tasks demand comprehensive graph-level representations. This crucial distinction emphasizes the indispensable role of the pooling mechanism in graph classification. The pooling mechanism is vital, since it efficiently transforms the input graph, enriched with node representations derived from GCNs, into a single vector or a size-reduced simplified graph. This transformation is crucial for capturing the graph's overall structure and characteristics, thereby facilitating more precise and insightful graph classification tasks.

Significant progress has been made in developing effective graph pooling methods, which are crucial for enhancing the performance of downstream tasks. Employing a hierarchical architecture, graph pooling captures node correlations, as highlighted in works such as~\cite{eigenpool, second-order, TAPool}. These methods can be broadly categorized into node clustering pooling and node drop pooling. Node clustering pooling methods (\textit{e.g.}, DiffPool~\cite{diffpool}, MinCutPool~\cite{mincut}, and StructPool~\cite{ structpool}) cluster nodes to form new ones, effectively preserving feature information. However, a major drawback of these methods is the distortion of the original graph structures. Furthermore, they require additional networks to learn a dense cluster assignment matrix, resulting in substantial computational and storage demands, especially for large graphs. On the other hand, node drop pooling methods (\textit{e.g.}, Graph U-Net~\cite{graph-u-net}, SAGPool~\cite{sagpool}, and GSAPool~\cite{gsapool}), focus on retaining the most representative nodes by evaluating their significance. This method effectively preserves crucial structural information and is more efficient and practical compared to node clustering pooling, especially for managing large-scale networks.

\begin{wrapfigure}[15]{r}{0.45\textwidth}
\vspace{-0.8cm}
\setlength{\abovecaptionskip}{-0.2cm}   
\begin{center}
\includegraphics[width=0.45\textwidth]{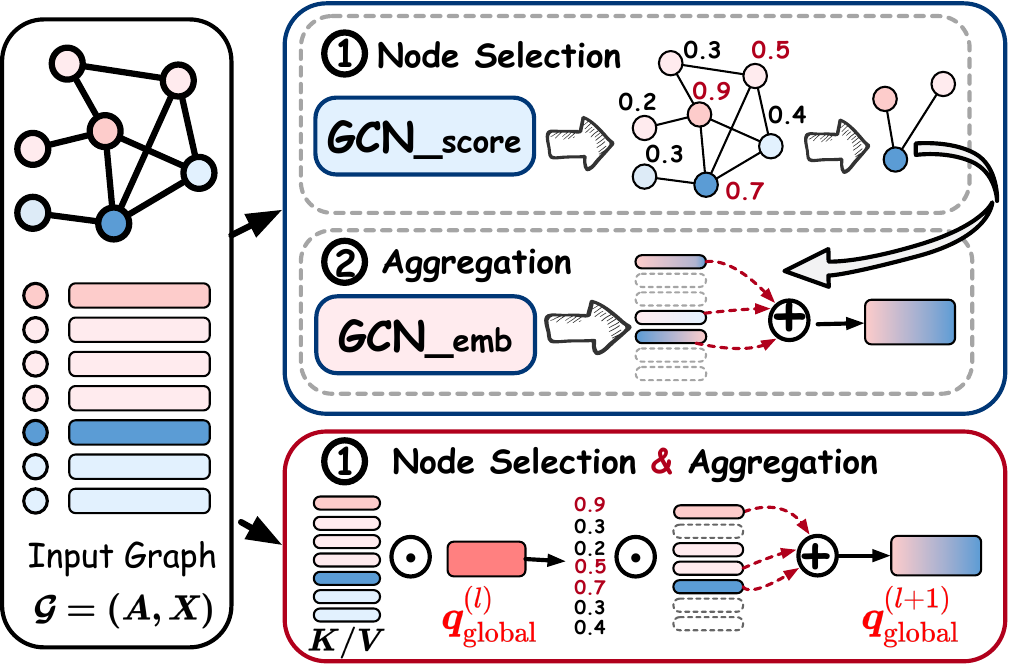}
\end{center}
\caption{Comparison between mainstream graph pooling methods (\textcolor{blue} {\textit{\textbf{top right}}}) and our proposed method (\textcolor{red} {\textit{\textbf{bottom right}}}). A detailed introduction to the symbols in the figure can be found in \textbf{$\S$\ref{sec:pre}}.}
\label{fig:motiv1}
\end{wrapfigure}

Although the node drop pooling method is renowned for its efficiency, it encounters challenges in mainstream applications. Node drop pooling iteratively discards nodes deemed less important, based on specific criteria, to achieve hierarchical representations. As illustrated in the top right of Figure~\ref{fig:motiv1}, prevalent pooling methods typically employ a single independent network to assign scores, indicating each node's significance. Although the scoring process is learnable, its \textbf{indirect connection to the final prediction} can sometimes cause sub-optimal node selection. Therefore, addressing this issue requires a graph pooling function that explicitly bases the retention of nodes on their contributions to the classification result. As depicted in the bottom right of Figure~\ref{fig:motiv1}, our proposed graph explicit pooling method (GrePool) selects nodes according to their impact on the final prediction results. Specifically, each node in the graph calculates attention scores in relation to an additional learnable global node, denoted as $\boldsymbol{q}_{\text{global}}$. Then, these attention scores are utilized to retain informative nodes. As a result, the capacity of pooling method can be flexibly controlled through the identification process where no additional parameters are introduced.  Consequently, the global node's embedding is created through a weighted combination of the retained nodes' embeddings based on their attention scores. Subsequently, the global node's embedding  serves as the basis for the final prediction. Hence, our method enables a strong correlation between the intertwined node selection and final prediction tasks, ensuring that the retained nodes truly contribute to the final prediction outcome. Compared to other node drop pooling models, GrePool takes into account the explicit influence on the classification result when performing node selection, without introducing additional parameters.


\begin{wrapfigure}[16]{r}{0.4\textwidth}
\vspace{-0.6cm}
\setlength{\abovecaptionskip}{-0.3cm}   
\begin{center}
\includegraphics[width=0.4\textwidth]{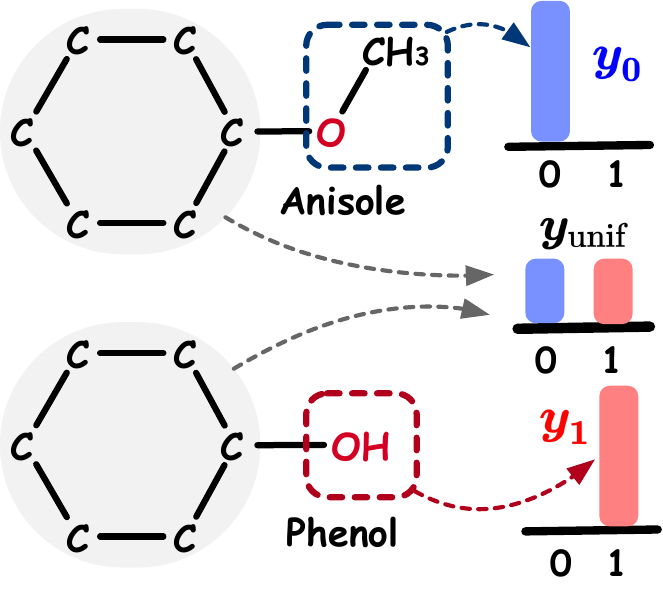}
\end{center}
\caption{Illustration of informative and uninformative nodes in the graph.}
\label{fig:motiv2}
\end{wrapfigure}

Current graph pooling methods typically prioritize informative nodes for information propagation and \textbf{neglect the discarded ones.} The discarded nodes may enhance performance; thus, it is necessary to re-examine these methods. Therefore, we propose an approach where uninformative nodes, which are unnecessary for classification in our view, are uniformly distributed across all categories instead of being completely dismissed (GrePool+). For example, in Figure~\ref{fig:motiv2}, Phenol and Anisole are organic compounds with a benzene ring. However, their chemical behavior is determined by their distinctive attributes: Phenol has a hydroxyl group (\ce{-OH}) while Anisole has a methoxy one (\ce{-CH3}). The presence of the hydroxyl group in Phenol makes it weakly acidic and highly reactive, whereas the methoxy group in Anisole makes it non-acidic and relatively reactive. These groups (hydroxyl and methoxy) are critical for classification since they provide important information. On the other hand, carbon rings are common elements in many organic compounds; therefore, they are less distinctive or informative. Given that these common patterns coexist across various categories, our approach aims to distribute their embeddings uniformly across all categories, applying uniform loss. This ensures equal prediction probabilities at any given category. This strategy's outcomes are two fold. First, it emphasizes the significance of the input graph's informative components while filtering out the trivial elements. Second, it facilitates the back-propagation of gradients through the uninformative nodes, ensuring a more comprehensive and balanced learning process.

In summary, this paper presents GrePool, an innovative attention-based pooling method that ensures the retained nodes explicitly contribute to the final prediction outcome. Moreover, we enhance this approach by applying uniform loss to the discarded nodes. This enhanced approach is called GrePool+. Furthermore, this refined approach emphasizes the identification of informative nodes, thereby improving the overall efficacy and precision of graph classification.  To evaluate our proposed method's effectiveness, we conduct extensive experiments on 12 commonly used datasets, including the large-scale Open Graph Benchmark (OGB)~\cite{ogb-dataset}. We also compare the results with 14 established baseline methods. The experimental results demonstrate that GrePool consistently surpasses the baseline methods for most of the datasets. Notably, the introduction of GrePool+ invariably boosts GrePool's performance without requiring additional computation. Our main contributions are summarized as follows: 
\begin{itemize}
  \item We propose an attention-based graph pooling method that explicitly selects reserved nodes based on their significant contribution to the final prediction outcome, concurrently eliminating the need for additional score-predicting networks commonly observed in conventional graph pooling methods. 
  \item We innovatively  harness the information from nodes that are commonly overlooked and discarded in conventional pooling methods, enhancing the training process and improving classification accuracy.
  \item Our proposed methods, GrePool and GrePool+, are consistently effective and generalizable across 12 widely used datasets, outperforming 14 baseline methods in extensive experimental evaluations.
\end{itemize}

\section{Preliminaries and Related Works}
\label{sec:pre}

\subsection{Notations}

$ G =(\mathcal{V}, \mathcal{E})$ denotes a graph with the node set $\mathcal{V}$ and edge set $\mathcal{E}$. The node attributes are denoted by $\boldsymbol{X} \in \mathbb{R}^{ n \times d}$, where $n$ is the number of nodes and $d$ is the node attribute dimension.  The graph topology is represented by an adjacency matrix $\boldsymbol{A} \in \{0, 1\}^{ n \times n}$. 

\subsection{Problem Statement}

\begin{definition}[\textbf{Graph Classification}]
\textit{Given a set of  graphs $\mathcal{D}=\left\{\left(G_1, y_1\right),\left(G_2, y_2\right), \cdots,\left(G_t, y_t\right)\right\}$, the primary objective of graph classification is to learn a mapping function $f$ that can effectively associate each input graph $G_i$ with its corresponding label $y_i$. This can be mathematically represented as:}
\begin{equation}
f : \mathcal{G} \rightarrow \mathcal{Y},
\label{eq:graph-classification}
\end{equation}
where $\mathcal{G}$ denotes the set of input graphs, $\mathcal{Y}$ represents the set of labels associated with the graphs, and $t$ signifies the total number of graphs in the dataset.
\end{definition}

\subsection{Graph Pooling}

\begin{definition}[\textbf{Graph Pooling}] 
\textit{A graph pooling operator, denoted as $\operatorname{POOL}$, is defined as any function that maps a given graph $ G =(\mathcal{V}, \mathcal{E})$ to a new pooled graph $G^{\prime}=(\mathcal{V}^{\prime}, \mathcal{E}^{\prime})$:}
\begin{equation}
G^{\prime}=\operatorname{POOL}(G),
\label{eq:pool}
\end{equation}
where $|\mathcal{V}^{\prime}| < |\mathcal{V}|$, and $|\mathcal{V}|$ represents the number of nodes in the original graph. It is worth noting that in certain exceptional cases, scenarios where $|\mathcal{V}^{\prime}| \geq |\mathcal{V}|$ may exist, resulting in upscaling the graph through pooling. The fundamental objective of graph pooling is to effectively reduce the number of nodes in a graph while capturing its hierarchical information.
\end{definition}

Graph pooling plays a crucial role in capturing the overall graph representation and can be broadly categorized into two types: global pooling and hierarchical pooling. Global pooling methods typically employ operations such as sum/average/max-pooling~\cite{duvenaud} or more sophisticated techniques~\cite{set2set,sortpool,deepsets,degree-pool,neural-readout} to aggregate node features and obtain graph-level representations. However, these methods often encounter information loss as they overlook the underlying graph structures.
To address this issue, hierarchical pooling models have been proposed, which are classified into node clustering pooling and node drop pooling. Node clustering pooling treats graph pooling as a clustering problem, where nodes are mapped into clusters as new nodes in a coarsened graph~\cite{eigenpool,diffpool,mincut,structpool,hoscpool,sep,c2n}. On the other hand, node drop pooling utilizes learnable scoring functions to identify a subset of nodes with lower significance scores from the original graph~\cite{TAPool, graph-u-net, sagpool,gsapool, ipool, asap,attpool, vip-pool, path-pooling, cgi-pool,MVpool,MGAP,MSGNN,Topopool}. Based on these selected nodes, a new coarsened graph is constructed by obtaining a new feature and adjacency matrix. Notably, node clustering pooling methods have limitations regarding storage complexity due to the computation of dense soft-assignment matrices. In contrast, node drop pooling methods are memory-efficient and better suited for large-scale graphs, although they may result in some information loss. For a more comprehensive understanding of these topics, we recommend referring to the recent reviews on graph pooling~\cite{pooling-review,understanding-pooling}, which provides in-depth insights into the various pooling methods.

\subsection{Attention in Graph Pooling}

\begin{definition}[\textbf{Graph Attention Mechanism}] 
\textit{The attention mechanism with various graph attention functions can be defined, within a generalized framework, as follows:}
\begin{equation}
\text{Attention} = p(q(\boldsymbol{X}), \boldsymbol{X}),
\end{equation}
where $q(\cdot)$ represents a function that generates attention to capture the node significance within the graph. The function $p(\cdot)$ utilizes the input data $\boldsymbol{X}$ to extract essential information based on the attention function. By processing the input data through the attention function, the model can extract relevant information, enhancing the overall performance and interpretability of the graph-based learning system.
\end{definition}

The attention mechanism has recently emerged as a powerful tool in natural language processing and computer vision. Its effectiveness in adaptively selecting discriminative features and filtering noise information has led to its integration into GNNs~\cite{gat}. Notably, attention mechanisms have also been introduced in graph pooling. One such approach is gPool~\cite{graph-u-net,understand-attention}, which employs a linear projection as an attention module to predict individual node coefficients. AttPool~\cite{attpool}, on the other hand, leverages local/global attention to select discriminative nodes and generate a graph representation through attention-weighted pooling. GMT~\cite{gmt} takes a different approach, utilizing multi-head attention~\cite{transformer} to compress the nodes into a small set of important nodes and calculate their inter-node relationships. In contrast to these existing methods, our proposed approach introduces a novel technique: multi-head self-attention. This technique enables us to perform node selection and information aggregation distinctively and effectively. By leveraging the power of self-attention, we can dynamically identify and prioritize the most relevant nodes in the graph, while simultaneously aggregating their information to generate a comprehensive representation.

\section{The Proposed Method}

This section provides a comprehensive analysis of the proposed method. We begin by introducing the key features and mechanisms of GrePool (\textbf{$\S$\ref{sec:GrePool}}) and its variant, GrePool+ (\textbf{$\S$\ref{sec:GrePool+}}). We then analyze GrePool to examine the power of its expressiveness (\textbf{$\S$\ref{sec:express}}). Next, we compare our method with several closely related approaches (\textbf{$\S$\ref{sec:discussion}}).  Finally, we explore the computational complexity of GrePool (\textbf{$\S$\ref{sec:further}}).

\begin{figure}[!t] 
\begin{center}
\includegraphics[width=0.95\linewidth]{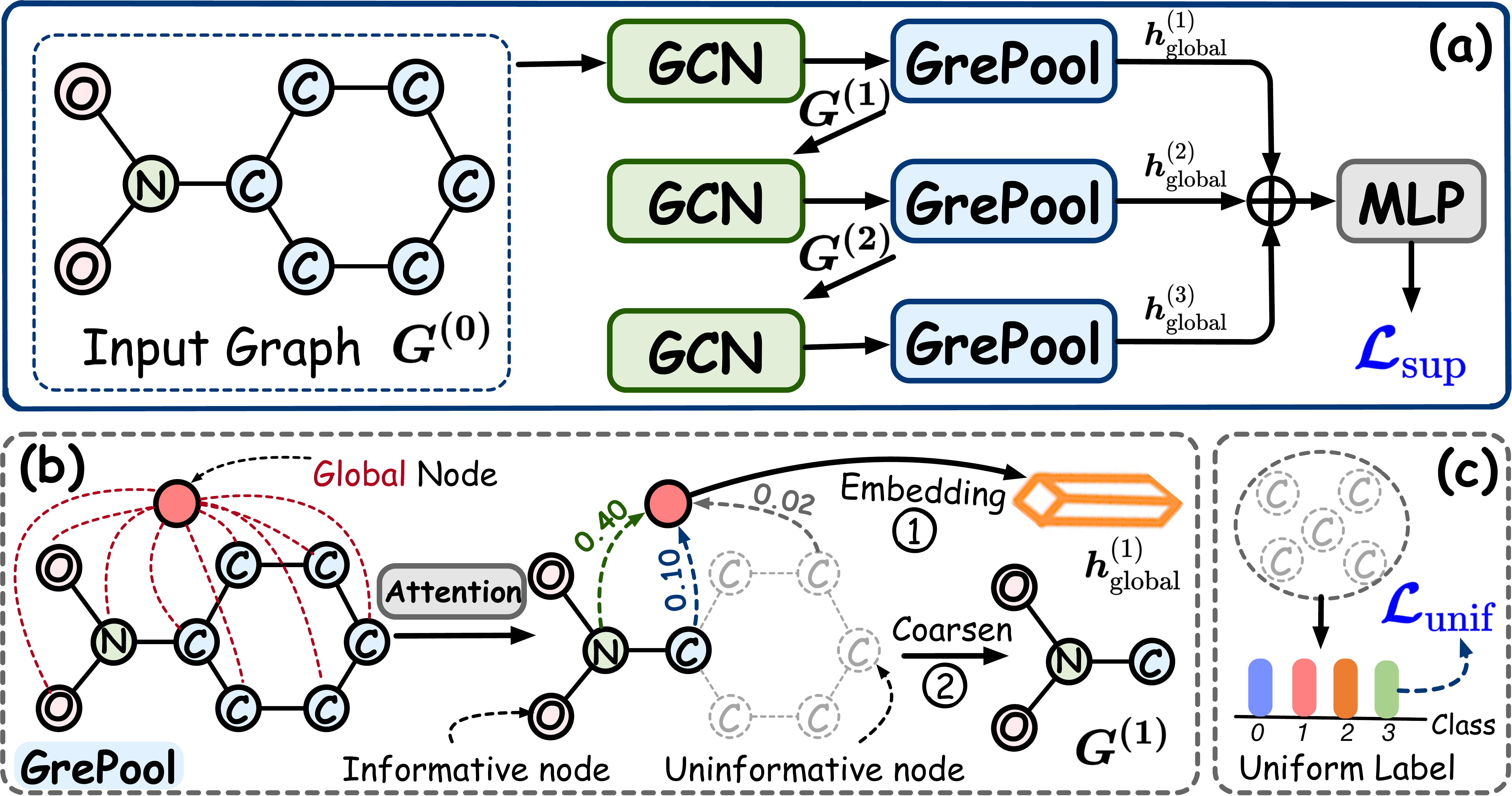}
\end{center}
\caption{An illustrative overview of our proposed method. \textbf{(a)} We present the overall graph pooling procedure, highlighting the integration of GrePool. \textbf{(b)} We explore the details of GrePool, emphasizing its ability to distinguish between informative and uninformative nodes based on the attention scores from the attention module. \textbf{(c)} We introduce the concept of uniform loss. This loss function assigns a uniform label to uninformative nodes' embeddings, facilitating their differentiation from informative nodes during training. The enhanced version of GrePool is referred to as GrePool+. }
\label{fig:model}
\end{figure}

\subsection{Graph Explicit Pooling (GrePool)}
\label{sec:GrePool}

This section introduces GrePool's key mechanisms, emphasizing its node selection and information aggregation  proficiency. As depicted in Figure~\ref{fig:model}, GrePool consists of three essential modules: \textbf{graph convolution}, \textbf{attention-based graph pooling}, and \textbf{optimization objective}. A detailed explanation of each module is presented below.

\paragraph{Graph Convolution} The GCN module is GrePool's fundamental building block, enabling the model to effectively capture and propagate information throughout the graph.  This module utilizes GCN to learn node representations by aggregating information from neighboring nodes. It plays a pivotal role in capturing the local structural patterns and inherent  features in the graph. A generic GNN layer can be expressed as follows:
\begin{equation}
\boldsymbol{h}_v^{(l)}=f_{(l)}\left(\boldsymbol{h}_v^{(l-1)},\left\{\boldsymbol{h}_u^{(l-1)} \mid u \in \mathcal{N}(v)\right\}\right),
\end{equation}
where $\mathcal{N}(v) \subseteq \mathcal{V}$ represents the neighborhood of node $v$, $\boldsymbol{h}_v^{(0)} = \boldsymbol{X} \in \mathbb{R}^{d}$ denotes the initial node representation, and $f_{(l)}(\cdot)$ is a function parameterized by a neural network. This function transforms and aggregates information from the previous to the current layer. Notably, this function can be incorporated into various GNN formulations.

\paragraph{Attention-based Graph Pooling}  In the preceding analysis (in Figure~\ref{fig:motiv2}), it is posited that only a subset of the node embeddings help to predict the labels of a graph for a given task, permitting the safe removal of other nodes without affecting the network output. Drawing inspiration from the application of Transformers in text classification~\cite{transformer}, we introduce a learnable global node, the output embedding of which encapsulates all pairwise interactions in a single classification vector. This approach offers two significant advantages: \textbf{1)} it improves aggregation methods, outperforming traditional, non-learned readout strategies such as sum/mean pooling; \textbf{2)} it reveals each node's contribution to the classification outcome by forwarding the global node to the classifier for label prediction, using the attention scores between the individual and global nodes. We expound on this module in the subsequent discourse.

The global node's embedding is updated using the self-attention mechanism, making the attention map a reflection of the relation or similarity metrics between the nodes in the graph and the global node, formally expressed as:
\begin{equation}
\boldsymbol{h}_{\text {global}}=\operatorname{softmax}\left(\frac{\boldsymbol{q}_{\text {global}} \cdot \boldsymbol{K}}{\sqrt{d}}\right) \boldsymbol{V}=\boldsymbol{a} \cdot \boldsymbol{V},
\end{equation}
where $\boldsymbol{q}_{\text {global}}, \boldsymbol{K},$ and $\boldsymbol{V}$ represent global node's query vectors, the key matrix, and the value matrix, respectively, within an attention head. Thus, the global node's output, $\boldsymbol{h}_{\text {global}}$, consists of a linear combination of the value vectors $\boldsymbol{V}=\left[\boldsymbol{h}_{\boldsymbol{v}_1}, \boldsymbol{h}_{\boldsymbol{v}_2}, \ldots, \boldsymbol{h}_{\boldsymbol{v}_n}\right]^{\top}$. The coefficients of this combination are the attention values corresponding to the global node in relation to all other nodes.

Given that $\boldsymbol{h}_{\boldsymbol{v}_i}$ is derived from the $i$-th node, the attention value $\boldsymbol{a}_i$ (the $i$-th element in $\boldsymbol{a}$) quantifies the extent to which information from the $i$-th node is integrated into the global node's output (\textit{i.e.}, $\boldsymbol{h}_{\text{global}}$) through linear combination. Therefore, inferring that the magnitude of $\boldsymbol{a}_i$ is indicative of the significance of the $i$-th node is reasonable. As a result, utilizing the attention values has become a straightforward and widely adopted method for interpreting model decisions. Hence, we leverage these attention scores to define the node importance score $\boldsymbol{S}$ for graph pooling during training and inference, facilitating the dynamic discrimination between informative and uninformative nodes within graphs. Specifically, after calculating the node significance scores in the graph, we perform a selection operation, which selects nodes with the highest $\lceil {p}\times {n} \rceil$ significance scores and coarsens the graph accordingly. In this instance, $p$ is the pooling ratio, similar to the established graph pooling methodologies~\cite{sagpool,graph-u-net}. This approach can be expressed as follows:
\begin{equation}
\mathrm{idx}^{(l)}=\operatorname{TOP}_{k}(\boldsymbol{S}^{(l)}); \quad 
\boldsymbol{X}^{(l+1)}=\boldsymbol{X}^{(l)}(\mathrm{idx}^{(l)},:) \odot \boldsymbol{S}^{(l)}(\mathrm{idx}^{(l)},:) ; \quad \boldsymbol{A}^{(l+1)}=\boldsymbol{A}^{(l)}(\mathrm{idx}^{(l)}, \mathrm{idx}^{(l)}),
\end{equation}
where $\operatorname{TOP}_{k}$ orders values and returns the top $k$ value indices in $\boldsymbol{S}^{(l)}$, and $\mathrm{idx}^{(l)}$ denotes the indices of the nodes retained for the subsequent graph in layer $l+1$.


\paragraph{The Optimization Objective}

We regard the global node embedding $\boldsymbol{h}_{\text{global}}$ from each network layer as a comprehensive representation of the entire graph. Then, the representation is subjected to a linear transformation and softmax activation, enabling prediction generation, as shown below:
\begin{equation}
\widehat{\mathbf{y}}=\operatorname{softmax}\left(\boldsymbol{W} ( \boldsymbol{h}_{\text{global}}^{(1)} + \boldsymbol{h}_{\text{global}}^{(2)} + \cdots + \boldsymbol{h}_{\text{global}}^{(L))} )\right),
\end{equation}
where $\boldsymbol{h}_{\text{global}}^{(l)}$ represents the global node embedding at the $l$-th layer, $\boldsymbol{W} \in \mathbb{R}^{d \times d^{\prime}}$  is the weight matrix, and $L$ denotes the total number of layers. Moreover, we aim to minimize the cross-entropy loss between the predictions and ground-truth graph labels to optimize our model. This is defined as:
\begin{equation}
\mathcal{L}_{\mathrm{sup}}=-\frac{1}{|\mathcal{D}|} \sum_{G \in \mathcal{D}} \mathbf{y}_{G}^{\top} \log \left(\widehat{\mathbf{y}}_{G}\right),
\end{equation}
where $\mathcal{L}_{\mathrm{sup}}$ represents the cross-entropy loss computed throughout the dataset $\mathcal{D}$, and $\mathbf{y}_{G}$ is the ground-truth label vector associated with the graph $G$.

\subsection{Graph Explicit Pooling with Uniform Loss (GrePool+)}
\label{sec:GrePool+}

Similar to other graph pooling methodologies, our GrePool method selectively drops nodes from the graph, specifically those with lower attention scores, as illustrated in Figure~\ref{fig:model} (b). However, concerns regarding the potential contributions of these dropped nodes to the prediction outcomes should be considered. In response, we apply a uniform loss to the discarded nodes to collect information from uninformative nodes, thereby enhancing our method.

More precisely, we hypothesize that nodes with lower attention scores converge at trivial patterns or nodes, which may be irrelevant to classification tasks. To counteract this, we employ an even distribution of the predictive probabilities of the above nodes across all categories. Thus, the uniform classification loss is defined as:
\begin{equation}
\mathcal{L}_{\text {unif}}=\frac{1}{|\mathcal{D}|} \sum_{G \in \mathcal{D}} \mathrm{KL}\left(\mathbf{y}_{\text {unif}}, \widetilde{\mathbf{y}}_{G}\right),
\end{equation}
where $\mathrm{KL}$ represents the Kullback-Leibler Divergence~\cite{kl-div}, $\widetilde{\mathbf{y}}_{G}$ denotes the discarded node embeddings, and  $\mathbf{y}_{\text{unif}}$ represents the uniform distribution across categories.  Subsequently, we define the comprehensive loss function for GrePool+ as:
\begin{equation}
\mathcal{L}_{\text {total}}= \mathcal{L}_{\text {sup}} + \lambda*\mathcal{L}_{\text {unif}},
\end{equation}
where $\lambda$ serves as a parameter for the trade-off between the primary objective ($\mathcal{L}_{\text {sup}}$) and the auxiliary uniform loss ($\mathcal{L}_{\text {unif}}$).  By optimizing the dual objectives, our approach successfully distinguishes between informative and uninformative nodes within the graph. Explicitly penalizing uninformative node embeddings encourages the GrePool method to prioritize and emphasize informative nodes, improving the quality of the resulting graph representation.




\subsection{Expressiveness Power of GrePool}
\label{sec:express}
In this section, we theoretically examine the GrePool methodology,  emphasizing its expressiveness capacity.  By leveraging the advancements made by powerful GNNs, we demonstrate that if our graph pooling function is injective, GrePool can achieve a level of expressiveness comparable to that of the renowned Weisfeiler–Lehman (WL) test~\cite{wl-test}. The WL test is widely acknowledged for its exceptional ability to distinguish the local structures within a graph.

\vspace{7pt}
\begin{theorem}
\textit{Let $\mathcal{A}: G \rightarrow \mathbb{R}^n$ denote a GNN adhering to the neighborhood aggregation paradigm and utilizing an attention-based aggregator in conjunction with a readout function. This suggests that $\mathcal{A}$ achieves its maximal discriminating ability— which involves distinguishing unique local structures and achieving the same level of discrimination as the 1-Weisfeiler-Lehman (1-WL) test for differentiating distinct global structures—when both aggregation and readout functions are designed to be injective.}
\end{theorem}
\vspace{7pt}

\begin{proof}
From Lemma 2 and Theorem 3 in~\cite{gin}, we know that when all functions in $\mathcal{A}$ are injective, $\mathcal{A}$ can reach the upper bound of its discriminating ability, which is the same as the WL test~\cite{wl-test}, when determining the graph isomorphism.  The detailed proofs can be found in~\cite{gin}.
\end{proof}

\vspace{7pt}
\begin{corollary}
\textit{ $\mathcal{F}$  is defined as the original attention-based aggregator and readout function. It operates on  a multi-set $H \in \mathcal{H}$ with $\mathcal{H}$ representing a node feature space that has been systematically transformed from the countable input feature space $\mathcal{X}$. When $\mathcal{F}$ is characterized by injectivity, it possesses the capability to map two disparate graphs, $G_{1}$ and $G_{2}$, onto distinct embedding spaces. This attribute ensures that the overarching process within the GrePool framework can achieve a level of expressiveness and discrimination analogous to that of the WL test. This injectivity is crucial for preserving the structural information's uniqueness during the transformation from the graph domain to the embedding space, thereby facilitating the effective discrimination between non-isomorphic graphs.}
\end{corollary}
\vspace{7pt}

\begin{proof} To streamline the proof, we explore the injectivity of the attention-based aggregator and readout function, and limit our discussion to graphs with a fixed number of nodes. We consider that each graph comprises $n$ nodes represented as matrices $\boldsymbol{X} \in \mathbb{R}^{n \times d}$, where $d$ denotes each node vector's dimension. The self-attention mechanism's transformations are defined as $ \boldsymbol{Q} = \boldsymbol{X}\boldsymbol{W}_Q, \boldsymbol{K} = \boldsymbol{X}\boldsymbol{W}_K, \boldsymbol{V} = \boldsymbol{X}\boldsymbol{W}_V,$
where $\boldsymbol{W}_Q, \boldsymbol{W}_K, \boldsymbol{W}_V$ are the weight matrices associated with queries, keys, and values, respectively. The self-attention output is then computed as: 
\begin{equation}
\text{Attention}(\boldsymbol{Q}, \boldsymbol{K}, \boldsymbol{V}) = \text{softmax}(\frac{\boldsymbol{Q}\boldsymbol{K}^T}{\sqrt{d}})\boldsymbol{V}.
\end{equation} 
In this instance, $ \boldsymbol{X}_1$ and $\boldsymbol{X}_2$ represent the node features of two distinct graphs. Our objective is to demonstrate that their respective self-attention outputs are uniquely distinguishable. Given the distinctness of $ \boldsymbol{X}_1$ and $\boldsymbol{X}_2$, their corresponding $\boldsymbol{Q}, \boldsymbol{K}, \boldsymbol{V}$ matrices will also be distinct, provided that the weight matrices $\boldsymbol{W}_Q, \boldsymbol{W}_K, \boldsymbol{W}_V$ are of full rank. This premise is grounded in the fact that multiplication by a full-rank matrix preserves the uniqueness of varying inputs. By examining the product $\boldsymbol{Q}\boldsymbol{K}^T$, we observe that the inputs $\boldsymbol{X}_1$ and $\boldsymbol{X}_2$ yield distinct matrices $\boldsymbol{Q}_1\boldsymbol{K}_1^T$ and $\boldsymbol{Q}_2\boldsymbol{K}_2^T$. Following the softmax operation, which acts on these unique matrices, the resulting distinct probability matrices—assuming no value overlap exists— produce unique output matrices when multiplied by $\boldsymbol{V}$.

Furthermore, the weighted readout can be approximated by any instance-wise feed-forward network, representing a transformative function $\phi: \mathbb{R}^d \rightarrow \mathbb{R}^{d^{\prime}}$. This function can be seamlessly constructed over multi-set elements $h \in H$, ensuring injectivity. Thus, assuming a fixed-node graph and a non-overlapping softmax function, the attention-based aggregator and readout function together constitute an injective function. Based on this injectivity, the overall architectural framework of our model exhibits a discrimination level equivalent to that of the WL test, affirming its efficacy in graph representation.

\end{proof}

\subsection{Discussion} 
\label{sec:discussion}

This section discusses the comparison of our proposed method with several closely related approaches. Through this analysis, we aim to highlight the distinctive features and advantages of our method and emphasize its unique contributions to graph representation learning.

\paragraph{GrePool vs. GMT} In comparison to GMT~\cite{gmt}, which utilizes multi-head attention to cluster the given graph into representative nodes and calculate the relationships between them, the GrePool method adopts multi-head attention to select informative nodes and summarize the global embedding using the attention scores. This allows GrePool to focus on capturing the most relevant and informative nodes, enhancing the discriminating ability of the resulting graph representation.

\paragraph{GrePool vs. CGIPool} On the other hand, CGIPool~\cite{cgi-pool}, introduces positive and negative coarsening modules with an attention mechanism to learn real and fake coarsened graphs. However, two primary distinctions exist between CGIPool and GrePool. First, CGIPool's positive and negative coarsening modules  maximize the mutual information between the input and coarsened graph using a discriminator. In contrast,  GrePool focuses on selecting informative nodes directly through multi-head self-attention. Second, while CGIPool adopts a GNN as the attention mechanism to generate attention scores, GrePool employs multi-head self-attention. This enables a more fine-grained analysis of the inter-node relationships and enhances the model's ability to capture complex dependencies within the graph structure.

\subsection{Complexity Analysis}
\label{sec:further}
The GrePool algorithm differs from existing pooling methods, such as SAGPool~\cite{sagpool}, by eliminating the need for a score prediction stage. This simplifies the process and avoids extra computational and parameter-related complexity. The only computation required is self-attention, which has a computational cost of $\mathcal{O}(n^2)$ in a single epoch, where $n$ represents the number of nodes in the graph. Real-world graphs typically comprise around 20-30 nodes, so they are relatively small. Furthermore, the core mechanism of our method, self-attention, is well-suited for efficient GPU-based matrix operations. Therefore, although some perceived increase in computational complexity exists, the actual process is highly efficient and suitable for real-world applications.

\subsection{Summary}

Our GrePool method utilizes self-attention mechanisms to intelligently select informative nodes within a graph. This node selection process is directly connected to the final prediction, allowing for more accurate and effective graph pooling. Importantly, our approach requires no additional parameters or significant computational overhead, distinguishing it from previous methods. Building upon the success of GrePool, our GrePool+ method utilizes the information from the dropped nodes,  typically ignored by previous graph pooling methods. By incorporating a uniform loss, we ensure that the dropped nodes contribute to the overall learning process, enhancing the model's ability to capture the full range of information within the graph.  To provide a comprehensive understanding of the capabilities of GrePool and GrePool+, we conduct a thorough analysis of their theoretical foundations and distinguishing factors. This comprehensive evaluation allows us to establish the significance and potential impact of our methods in the field of graph representation learning.



\section{Experiment}
\label{sec:experiment}

\subsection{Experimental Settings}
\label{sec:exper-seeting}

\paragraph{Datasets} We comprehensively evaluated our method using 12 graph datasets. These datasets consist of six biochemical, two social from TU Datasests~\cite{tu-dataset}, and four large-scale datasets from the Open Graph Benchmark (OGB)~\cite{ogb-dataset}. Including these real-world datasets provides a wide-range of content domains and dataset sizes for a robust assessment of our method's performance. A clear overview of the dataset characteristics is provided in Tables~\ref{tab:result-tu} and~\ref{tab:result-ogb}. 

\paragraph{Models} To validate the superiority of GrePool, we used 14 methods as baselines for a comprehensive comparison:  \textbf{1) GNN-based methods} such as GCN~\cite{gcn} and GIN~\cite{gin};  \textbf{2) Flat pooling methods} such as Set2set~\cite{set2set} and SortPool~\cite{sortpool};  \textbf{3) Node clustering pooling methods}, including DiffPool~\cite{diffpool}, MinCutPool~\cite{mincut}, MemPool~\cite{mem-pool}, HaarPool~\cite{haar-pooling}, and GMT~\cite{gmt};  \textbf{4) Node drop pooling methods}, including TopKPool~\cite{graph-u-net}, SAGPool~\cite{sagpool}, GSAPool~\cite{gsapool}, and ASAP~\cite{asap}; \textbf{5) Edge-based pooling method} such as EdgePool~\cite{edgepool}.
The diverse range of baseline methodologies ensures that our comparative assessment is robust, spanning various approaches and paradigms in graph pooling.

\begin{table*}[!t]
\setlength{\belowcaptionskip}{0.3cm}
\renewcommand\arraystretch{1.3} 
\setlength\tabcolsep{2pt} 
\centering
\caption{Performance across nine datasets in the graph classification task. The reported results are mean and standard deviations over 10 different runs. {\color[HTML]{AE002F} \textbf{1) Red}:} the\textbf{ best} performance per dataset.  {\color[HTML]{0A6D6C} \textbf{2) Green}:} the \textbf{second best} performance per dataset. {\color[HTML]{3333FF} \textbf{3) Blue:}} the \textbf{third best} performance per dataset.}
\label{tab:result-tu}

\resizebox{\textwidth}{!}{%
\begin{tabular}{@{}lccccccccc@{}}
\toprule

\textbf{} &
  \multicolumn{6}{c}{\textbf{  Biochemical Domain (6)}} & \multicolumn{2}{c}{\textbf{\normalsize  Social Domain (2)}}
   \\ \cmidrule(lr){2-7}  \cmidrule(lr){8-9}
 &
  \textbf{NCI1} &
  \textbf{MUTAG} &
  \textbf{PTC-MR} &
  \textbf{NCI109} &
  \textbf{ENZYMES} &
  \textbf{MUTAGE.} & 
\textbf{IMDB-M} &
\textbf{COLLAB}
  \\ \midrule
  \# graphs &
  4,110 &
  188 &
  344 &
  4,127 &
  600 &
  4,337 & 
  1,500 &
  5,000
   \\ 
 \# nodes &
  29.87 &
  17.93 &
  14.29 &
  29.68 &
  32.63 &
  30.32 & 
  13.00 &
  74.49
   \\  
   \midrule
   

GCN &
  $76.03_{\pm 3.52}$ &
  $67.00_{\pm 11.4}$ &
  $52.57_{\pm 8.30}$ &
  $76.45_{\pm 2.34}$ &
  $44.84_{\pm 9.10}$ &
  $78.64_{\pm 1.60}$ &
  $50.40_{\pm 3.12}$ &
  $79.22_{\pm 1.59}$\\

GIN &
  $70.44_{\pm 2.49}$ &
  $64.00_{\pm 8.89}$ &
  $53.71_{\pm 8.11}$ &
  $69.88_{\pm 2.06}$ &
  $43.33_{\pm 6.41}$ &
  $75.47_{\pm 2.24}$ &
  $49.93_{\pm 2.92}$ &
  $78.52_{\pm 2.01}$\\ \midrule

Set2set &
  $71.00_{\pm 4.14}$ &
  $67.50_{\pm 6.02}$ &
  $54.86_{\pm 9.63}$ &
  $69.40_{\pm 3.00}$ &
  $44.00_{\pm 8.44}$ &
  $80.11_{\pm 1.73}$ &
  $50.47_{\pm 2.78}$ &
  $78.54_{\pm 1.74}$
   \\
SortPool &
  $74.14_{\pm 2.66}$ &
  $78.56_{\pm 10.5}$ &
  $56.00_{\pm 9.23}$ &
  $72.85_{\pm 2.26}$ &
  $31.50_{\pm 9.76}$ &
  $77.93_{\pm 1.90}$ &
  $50.20_{\pm 2.98}$ &
  $78.90_{\pm 1.79}$\\
  EdgePool &
  $76.53_{\pm 0.50}$ &
  $73.00_{\pm 0.87}$ &
  $54.29_{\pm 4.67}$ &
  $75.56_{\pm 2.53}$ &
  $36.17_{\pm 12.7}$ &
  $81.71_{\pm 1.92}$ &
  $49.80_{\pm 2.51}$ &
  $81.20_{\pm 1.42}$ \\
DiffPool &
  $77.04_{\pm 0.73}$ &
  $82.50_{\pm 2.54}$ &
  $55.26_{\pm 3.84}$ &
  $75.38_{\pm 0.66}$ &
  ${\color[HTML]{3333FF}\textbf{51.27}}_{\pm 2.89}$ &
  $79.80_{\pm 0.24}$ &
  ${\color[HTML]{AE002F}\textbf{51.03}}_{\pm 0.48}$ &
  $79.24_{\pm 1.99}$\\
MinCutPool &
  $75.26_{\pm 2.57}$ &
  $81.00_{\pm 10.2}$ &
  $55.43_{\pm 3.54}$ &
  $73.99_{\pm 1.82}$ &
  $48.83_{\pm 12.5}$ &
  $78.48_{\pm 2.29}$ &
  $50.33_{\pm 3.63}$ &
  ${\color[HTML]{3333FF}\textbf{81.16}}_{\pm 1.25}$\\
 HaarPool &
  $77.06_{\pm 1.91}$ &
  $62.50_{\pm 2.50}$ &
  ${\color[HTML]{3333FF}\textbf{58.14}}_{\pm 7.98}$ &
  $70.97_{\pm 1.89}$ &
  $43.67_{\pm 5.26}$ &
  $79.36_{\pm 2.13}$ &
  $47.73_{\pm 2.24}$ &
  $80.66_{\pm 1.59}$ \\
  MemPool &
  $63.43_{\pm 2.78}$ &
  $63.50_{\pm 8.67}$ &
  $53.43_{\pm 9.90}$ &
  $62.50_{\pm 2.37}$ &
  $46.83_{\pm 7.80}$ &
  $72.02_{\pm 2.27}$ &
  $48.13_{\pm 3.07}$ &
  $78.56_{\pm 1.42}$\\
GMT &
  ${\color[HTML]{3333FF}\textbf{77.32}}_{\pm 1.97}$ &
  $83.00_{\pm 11.8}$&
  $55.71_{\pm 9.54}$ &
  ${\color[HTML]{3333FF}\textbf{78.04}}_{\pm 1.88}$ &
  $48.29_{\pm 7.12}$ &
  ${\color[HTML]{3333FF}\textbf{81.95}}_{\pm 1.73}$ &
  $50.40_{\pm 1.83}$ &
  $80.62_{\pm 1.96}$\\ 
SAGPool &
  $71.95_{\pm 2.81}$ &
  $67.50_{\pm 7.83}$ &
  $55.43_{\pm 7.47}$ &
  $70.99_{\pm 2.37}$ &
  $42.67_{\pm 8.92}$ &
  $76.23_{\pm 3.73}$ &
  $49.87_{\pm 3.51}$ &
  $79.64_{\pm 2.19}$\\

TopKPool &
  $73.48_{\pm 2.07}$ &
  ${\color[HTML]{3333FF}\textbf{84.50}}_{\pm 9.34}$ &
  $54.57_{\pm 10.3}$ &
  $73.07_{\pm 1.87}$ &
  $40.83_{\pm 7.97}$ &
  $75.68_{\pm 4.65}$ &
  $50.20_{\pm 2.98}$ &
  $78.56_{\pm 2.11}$\\

GSAPool &
  $73.31_{\pm 3.70}$ &
  $63.00_{\pm 10.5}$ &
  $53.59_{\pm 2.69}$ &
  $72.25_{\pm 2.24}$ &
  $45.67_{\pm 9.58}$ &
  $77.98_{\pm 3.25}$ &
  $49.80_{\pm 3.46}$ &
  $78.64_{\pm 1.78}$\\
ASAP &
  $74.82_{\pm 2.90}$ &
  $73.00_{\pm 11.6}$ &
  $55.24_{\pm 4.86}$ &
  $72.20_{\pm 3.18}$ &
  $42.33_{\pm 7.79}$ &
  $80.16_{\pm 2.13}$ &
  $49.60_{\pm 3.23}$ &
  $--$\\\midrule

GrePool &
  ${\color[HTML]{0A6D6C}\textbf{82.62}}_{\pm 2.21}$ &
  ${\color[HTML]{0A6D6C}\textbf{86.25}}_{\pm 8.35}$ &
  ${\color[HTML]{0A6D6C}\textbf{59.86}}_{\pm 6.67}$ &
  ${\color[HTML]{0A6D6C}\textbf{82.13}}_{\pm 1.57}$ &
  ${\color[HTML]{0A6D6C}\textbf{51.92}}_{\pm 5.64}$ &
  ${\color[HTML]{0A6D6C}\textbf{83.03}}_{\pm 1.79}$ &
  ${\color[HTML]{3333FF}\textbf{50.77}}_{\pm 3.25}$ &
  ${\color[HTML]{0A6D6C}\textbf{81.42}}_{\pm 1.53}$\\ 

GrePool+ &
  ${\color[HTML]{AE002F}\textbf{83.07}}_{\pm 1.73}$ &
  ${\color[HTML]{AE002F}\textbf{88.50}}_{\pm 6.73}$ &
  ${\color[HTML]{AE002F}\textbf{62.86}}_{\pm 6.70}$ &
  ${\color[HTML]{AE002F}\textbf{82.15}}_{\pm 1.92}$ &
  ${\color[HTML]{AE002F}\textbf{52.92}}_{\pm 6.81}$ &
  ${\color[HTML]{AE002F}\textbf{83.30}}_{\pm 1.71}$ &
  ${\color[HTML]{AE002F}\textbf{51.10}}_{\pm 2.87}$&
  ${\color[HTML]{AE002F}\textbf{81.51}}_{\pm 1.19}$
 
 \\ \bottomrule
\end{tabular}%
}
\end{table*}

\begin{figure}[!t] 
\begin{center}
\includegraphics[width=1.0\linewidth]{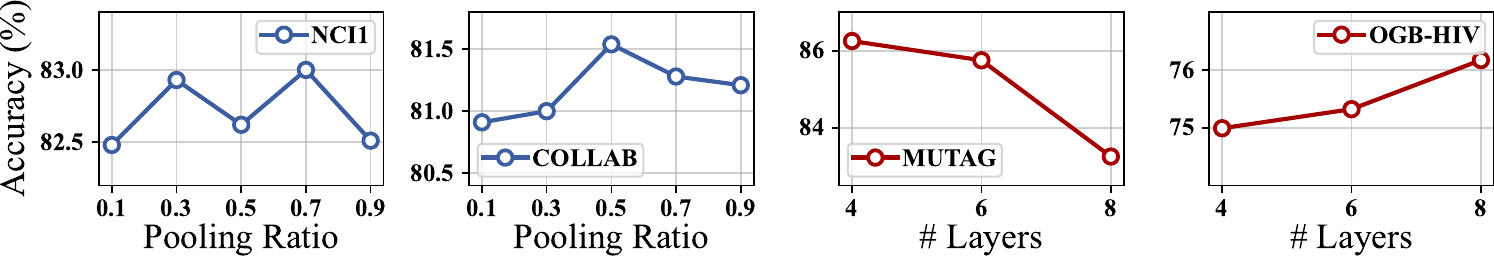}
\end{center}
\caption{Performance of GrePool with distinct pooling ratios and a different number of layers.}
\label{fig:para-intra}
\end{figure}

\paragraph{Implementation Details} To ensure a fair comparison, we standardized the pooling ratio to 0.5 and 0.25 for the TU and OGB datasets, respectively, across all methods, following the established settings outlined in~\cite{gin,gmt}. Additionally, we adopted the parameter settings (excluding the pooling ratio) specified in the corresponding papers for certain comparative models. In cases where the parameter settings were not provided, we conducted parameter tuning to optimize the model's performance.
Accuracy was selected as the metric and 10 runs were performed to ensure the reliability of the results. Furthermore, we explored the additional parameter $\lambda$ within the range of $\{0.01, 0.1, 1\}$ for our method. 

\subsection{Experimental Results} 
The main experimental results are presented in Tables~\ref{tab:result-tu} and~\ref{tab:result-ogb}, which provide valuable insights into our method's performance. Through a thorough analysis of these results, we uncovered several significant and insightful findings. In the following sections, we will explore these findings in detail.

\paragraph{Performance of GrePool} \underline{First},  GrePool consistently surpasses competing methods on nearly all examined datasets, adequately proving its efficacy. This validation reinforces the superiority of our approach in achieving superior performance in graph classification tasks. \underline{Second}, in comparison to GCN-based and flat pooling methods such as GCN, GIN, Set2Set, and SortPool, the GrePool method exhibits substantial improvements on most datasets.

\begin{wraptable}[23]{r}{0.50\textwidth}
\vspace{-0.4cm}
\centering
\setlength{\belowcaptionskip}{0.3cm}
\renewcommand\arraystretch{1.3} 
\setlength\tabcolsep{2pt} 
\caption{Performance of graph classification task on four OGB datasets. }
\label{tab:result-ogb}
\resizebox{0.5\textwidth}{!}{%
\begin{tabular}{@{}lcccc@{}}
\toprule
           & \multicolumn{4}{c}{\textbf{OGB Datasets (4)}}                                                  \\ \midrule
           & \textbf{HIV}                 & \textbf{BBPB}                & \textbf{TOX21}              & \textbf{TOXCAST}             \\ \midrule
\# graphs    & 41,127 & 2,039 & 7,831 & 8,576 \\
\# nodes   & 25.51 & 24.06 & 18.57 & 18.78 \\  \midrule
Set2Set    & $73.42_{ \pm 2.34}$ & $64.43_{ \pm 2.16}$ & $73.42_{ \pm 0.67}$ & $59.76_{ \pm 0.65}$ \\
SortPool   & $71.88_{ \pm 1.83}$ & $64.33_{ \pm 3.10}$ & $68.90_{ \pm 0.78}$ & $59.28_{ \pm 0.99}$ \\
EdgePool   & $72.15_{ \pm 1.56}$ & ${\color[HTML]{AE002F}\textbf{68.56}}_{ \pm 1.43}$ & $74.54_{ \pm 0.79}$ & $62.57_{ \pm 1.36}$ \\
DiffPool   & $75.05_{ \pm 1.71}$ & $64.77_{ \pm 2.43}$ & $75.82_{ \pm 0.69}$ & ${\color[HTML]{3333FF}\textbf{65.79}}_{ \pm 0.87}$ \\
MinCutPool & $73.91_{ \pm 1.10}$ & $66.47_{ \pm 1.90}$ & ${\color[HTML]{AE002F}\textbf{78.78}}_{ \pm 0.61}$ & $63.66_{ \pm 1.56}$ \\
MemPool    & $73.75_{ \pm 1.90}$ & $66.47_{ \pm 1.90}$ & $72.05_{ \pm 0.93}$ & $61.85_{ \pm 0.36}$ \\
GMT        & ${\color[HTML]{AE002F}\textbf{76.41}}_{ \pm 2.32}$ & ${\color[HTML]{3333FF}\textbf{66.88}}_{ \pm 1.59}$ & $76.56_{ \pm 0.90}$ & $64.53_{ \pm 0.92}$ \\
SAGPool    & $70.19_{ \pm 3.66}$ & $64.29_{ \pm 2.96}$ & $69.39_{ \pm 1.88}$ & $59.09_{ \pm 1.38}$ \\
TopKPool   & $71.24_{ \pm 2.97}$ & $65.93_{ \pm 2.60}$ & $68.69_{ \pm 2.02}$ & $58.63_{ \pm 1.56}$ \\
GSAPool    & $71.47_{ \pm 2.43}$ & $64.49_{ \pm 3.31}$ & $69.18_{ \pm 2.05}$ & $59.60_{ \pm 1.17}$ \\
ASAP       & $71.60_{ \pm 1.71}$ & $61.93_{ \pm 3.18}$ & $70.00_{ \pm 1.50}$ & $60.32_{ \pm 1.34}$ \\\midrule
GrePool    & ${\color[HTML]{0A6D6C}\textbf{76.17}}_{ \pm 1.06}$ & $66.21_{ \pm 1.69}$ & ${\color[HTML]{3333FF}\textbf{77.27}}_{ \pm 0.54}$ & $ {\color[HTML]{0A6D6C}\textbf{65.92}}_{ \pm 0.75}$ \\ 
GrePool+    & ${\color[HTML]{3333FF}\textbf{75.82}}_{ \pm 1.48 }$ & ${\color[HTML]{0A6D6C}\textbf{66.91}}_{ \pm 1.74}$ & ${\color[HTML]{0A6D6C}\textbf{77.62}}_{ \pm 0.44}$ & ${\color[HTML]{AE002F}\textbf{65.97}}_{ \pm 0.58}$ \\  \bottomrule
\end{tabular}%
}
\end{wraptable}
This is because  GCN-based and flat pooling methods do not consider the graphs' hierarchical structures, a limitation that our method effectively addresses. Moreover, these findings highlight the importance of incorporating hierarchical pooling layers in graph representation learning. \underline{Third}, the GrePool method outperforms node drop pooling methods such as SAGPool, indicating a more efficient strategy in retaining nodes critical for performance. \underline{Fourth},
as indicated in Table~\ref{tab:result-tu}, the GrePool method demonstrates increasingly pronounced enhancements over the baseline models as the dataset size rises. For instance, on the NCI1, NCI109, and MUTAGENICITY datasets, GrePool's performance improved by $6.85\%$, $7.42\%$, and $2.87\%$, respectively. These significant improvements indicate GrePool's scalability, especially in the context of large datasets. Notably, due to memory and computational resource limitations, some memory-intensive or time-consuming pooling baselines, such as HaarPool, are excluded from the results in Table~\ref{tab:result-ogb} for the OGB datasets. 


\paragraph{Performance of GrePool+}  The GrePool method utilizing uniform loss on uninformative nodes consistently outperforms the alternative that does not apply uniform loss. Although the improvement may seem modest, it is achieved without introducing any additional computational overhead. This demonstrates that preserving informative nodes while retaining the information from the uninformative ones is more effective than only keeping informative nodes. This also highlights the effectiveness of our informative node identification strategy, since the majority of informative nodes are well preserved. Furthermore, we note that GrePool+ displays smaller accuracy fluctuations in accuracy compared to the approaches that discard uninformative nodes. This is evident in the lower standard deviation values presented in Tables~\ref{tab:result-tu} and~\ref{tab:result-ogb} for MUTAG, NCI1, and TOXCAST datasets. The reduced fluctuation indicates that the including uniform loss on informative nodes improves accuracy and enhances training stability.



\subsection{Parameter Analysis}

\begin{wrapfigure}[11]{r}{0.5\textwidth}
\vspace{-0.6cm}
\begin{center}
\includegraphics[width=0.5\textwidth]{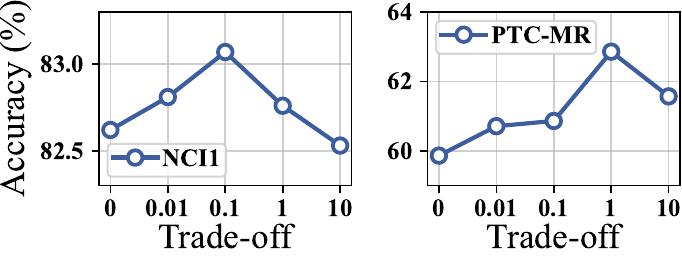}
\end{center}
\caption{Performance of GrePool with different trade-off parameters on two graph datasets.}
\label{fig:para-inter}
\end{wrapfigure} 

\paragraph{Impact of Pooling Ratio, Number of Layers and  Trade-off Parameter} We conducted an in-depth analysis of the effects of $L$, $p$, and $\lambda$ using GrePool on five graph datasets: NCI1, COLLAB, MUTAG, PTC-MR, and OGB-HIV. The comprehensive results of our analysis are presented in Figures~\ref{fig:para-intra} and~\ref{fig:para-inter}.  \underline{First}, we investigated the impact of the pooling ratio on graph classification performance. Our findings reveal that employing large pooling ratios result in performance fluctuations, indicating the presence of redundant information within the graphs. 
Specifically, larger pooling ratios introduce an increased amount of redundant information, which can hinder performance rather than enhance it. Furthermore, GrePool's accuracy range  is relatively small, suggesting that our method effectively selects essential nodes for graph-level representation learning, regardless of the pooling ratio. \underline{Second}, we examined the effect of increasing the value of $L$. Our observations demonstrate that for small-scale datasets such as MUTAG, the test accuracy decreases as $L$ increases. Conversely, for relatively large-scale datasets like OGB-HIV, the test accuracy exhibits an upward trend. Such phenomenon can be attributed to the potential overfitting of deeper GrePool models when applied to small-scale datasets. \underline{Third}, we explored the impact of the trade-off parameter $\lambda$. The results indicate that our model performs optimally when $\lambda$ is set around 0.1 and 1. Values that are too large or small can have a detrimental effect on the model's performance.

\begin{wrapfigure}[11]{r}{0.5\textwidth}
\vspace{-0.6cm}
\begin{center}
\includegraphics[width=0.5\textwidth]{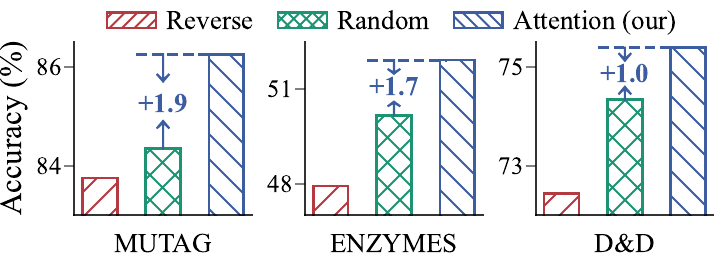}
\end{center}
\caption{Model performance with different node selection strategies. }
\label{fig:strategy}
\end{wrapfigure}

\paragraph{Impact of Node Selection Strategies}
In this section, we aim to evaluate the effectiveness of our proposed method by testing the effects of different node selection strategies on various graph datasets.  Specifically, we compared our attention-based node selection strategy with random and reverse selection strategies, as shown in Figure~\ref{fig:strategy}. Random selection refers to randomly selecting informative tokens, whereas reverse selection involves selecting nodes with the lowest attention scores in the GrePool model.  
All the node selection strategies were conducted  under identical settings, except for the node selection process. The results in Figure~\ref{fig:strategy} demonstrate that our attention-based node selection strategy outperforms the other selection strategies in accuracy. Additionally, the reverse selection strategy's performance underperforms compared to random selection, indicating that our method effectively identifies essential nodes for graph-level representation learning, while the discarded nodes are deemed uninformative for the classification task.  Furthermore, it is observed that with the increase in the average number of nodes in the graphs (\textit{e.g.}, MUTAG: 17.9, ENZYMES: 32.63, and D\&D: 284.3), the performance gap between GrePool and random selection decreases. This observation suggests that the selection strategy is essential for graphs with fewer nodes. It is notable that the average node number in most real-world datasets for graph classification tasks typically ranges from 20 to 30.

\begin{wrapfigure}[14]{r}{0.5\textwidth}
\vspace{-0.5cm}
\begin{center}
\includegraphics[width=0.5\textwidth]{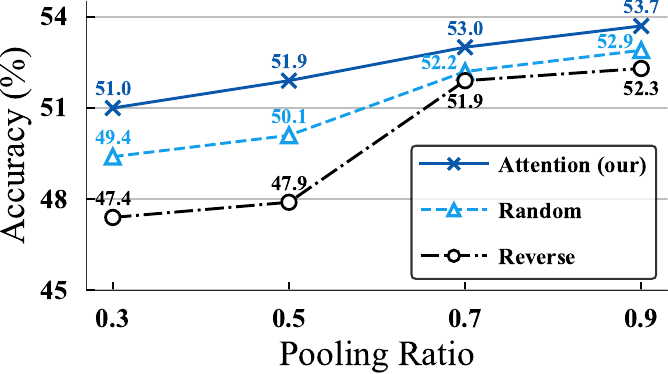}
\end{center}
\caption{Performance of different strategies in different pooling ratios.}
\label{fig:strategy2}
\end{wrapfigure}

Furthermore, we investigated the performance of different strategies with varying pooling ratios. The results presented in Figure~\ref{fig:strategy2} demonstrate that as the pooling ratio increases, the performance gap among the three strategies becomes narrower. Notably, when the pooling ratio reaches 0.9, random selection's accuracy nearly equals to our attention-based method. This is because the selection strategy becomes less influential as the number of remaining nodes increases. Actually,  as shown in Figure~\ref{fig:strategy2}, pooling methods mostly perform better when the pooling ratio is around from 0.5 to 0.7, since redundant information is present in the graphs.

\subsection{Broader Evolution}

\begin{wraptable}[10]{r}{0.50\textwidth}
\vspace{-0.4cm}
\caption{Results of graph classification task with two baseline methods on five different size datasets. }
\label{tab: broader-evolution}
\centering
\renewcommand\arraystretch{1.3} 
\setlength\tabcolsep{1pt} 
\resizebox{0.5\textwidth}{!}{%
\begin{tabular}{@{}lccccc@{}}
\toprule
         &\textbf{ MUTAG }       & \textbf{PTC-MR}       & \textbf{NCI109}      & \textbf{FRAN.} &\textbf{COLLAB}       \\ \midrule
SAGPool  & $67.50_{\pm 7.83}$ & $55.43_{\pm 7.47}$ & $\textbf{70.99}_{\pm  2.37}$ & $59.70_{\pm 2.54}$ & $79.64_{\pm  2.19}$ \\
SAGPool+ &  $\textbf{71.00}_{\pm 6.15}$ & $\textbf{56.59}_{\pm 5.81}$ & $70.44_{\pm 2.43}$ & $\textbf{60.20}_{\pm 2.48}$ &  $\textbf{80.97}_{\pm 1.86}$ \\
\textbf{Impro.}   & \cellcolor[HTML]{FDECEE} 5.19\% $\bm{\uparrow}$      & \cellcolor[HTML]{FDECEE} 2.09\% $\bm{\uparrow}$      & \cellcolor[HTML]{E5F2FC} 0.7\% $\bm{\downarrow}$       & \cellcolor[HTML]{EBEBEB} 0.8\%  $\bm{\uparrow}$      & \cellcolor[HTML]{FDECEE} 1.67\% $\bm{\uparrow}$      \\ \midrule
GSAPool  & $63.00_{\pm 10.5}$  & $53.59_{\pm 2.69}$ & $72.25_{\pm 2.24}$ & $60.30_{\pm 2.66}$  & $78.64_{\pm 1.78}$  \\
GSAPool+ & $\textbf{67.50}_{\pm 9.69}$ & $\textbf{56.18}_{\pm 1.43}$  & $\textbf{73.95}_{\pm 1.81}$  & $\textbf{60.48}_{\pm 2.48}$ & $\textbf{79.44}_{\pm1.30}$ \\
\textbf{Impro.}   &\cellcolor[HTML]{FDECEE} 7.14\% $\bm{\uparrow}$      & \cellcolor[HTML]{FDECEE} 4.83\%  $\bm{\uparrow}$     &\cellcolor[HTML]{FDECEE}  2.35\% $\bm{\uparrow}$      &\cellcolor[HTML]{EBEBEB}  0.3\%  $\bm{\uparrow}$      & \cellcolor[HTML]{FDECEE} 1.0\% $\bm{\uparrow}$       \\ \bottomrule
\end{tabular}%
}
\end{wraptable}
\paragraph{Evaluation on Other Graph Pooling Methods} 
To evaluate the generalization ability of the uniform loss operation, we extended its application to other node drop pooling methods.  
Specifically, we selected two representative pooling methods, namely SAGPool~\cite{sagpool} and GSAPool~\cite{gsapool}, and conducted experiments on five diverse graph datasets with varying sizes and domains. The experimental settings were kept consistent with those employed in the GrePool experiments. As illustrated in Table~\ref{tab: broader-evolution}, SAGPool+ and GSAPool+ denote the integration of uniform loss with SAGPool and GSAPool respectively. The results exhibit a marked improvement in performance attributable to our methodological enhancements, with pronounced benefits observed in small-scale graphs. This empirical evidence aligns with the advancements observed in the GrePool+ experiments previously discussed, further validating the efficacy and adaptability of our proposed approach.

\begin{wraptable}[8]{r}{0.50\textwidth}
\vspace{-0.05cm}
\centering
\renewcommand\arraystretch{1.6} 
\setlength\tabcolsep{1pt} 
\caption{Results of node classification across six datasets. The reported results are mean and standard deviation over 10  runs.}
\label{tab:nc}
\resizebox{0.5\textwidth}{!}{%
\begin{tabular}{@{}lcccccc@{}}
\toprule
\textbf{}     & \textbf{Cornell}   & \textbf{Texas}       & \textbf{Wiscon.}   & \textbf{Actor}      & \textbf{Squirrel}    & \textbf{Chamel.}   \\ \midrule
GAT           & $43.51_{\pm 7.1}$         & $58.65_{\pm 4.2}$          & $52.35_{\pm2.6}$          & $27.57_{\pm 0.8}$         & $27.29_{\pm 1.6}$          & $43.55_{\pm 2.2}$          \\
GAT+ & $\textbf{48.92}_{\pm 5.2}$ &  $\textbf{62.97}_{\pm 5.8}$ & $\textbf{54.51}_{\pm 4.0}$ & $\textbf{28.30}_{\pm 1.0}$ & $\textbf{27.5}0_{\pm 1.4}$ & $\textbf{43.86}_{\pm 2.3}$ \\ \midrule
\textbf{Impro.}        & \cellcolor[HTML]{FDECEE} 12.4\% $\bm{\uparrow}$            & \cellcolor[HTML]{FDECEE} 7.36\% $\bm{\uparrow}$               & \cellcolor[HTML]{FDECEE} 4.1\% $\bm{\uparrow}$               & \cellcolor[HTML]{FDECEE} 2.6\% $\bm{\uparrow}$              &  \cellcolor[HTML]{EBEBEB} 0.7\%   $\bm{\uparrow}$              & \cellcolor[HTML]{EBEBEB} 0.7\% $\bm{\uparrow}$               \\ \bottomrule
\end{tabular}%
}
\end{wraptable}

\paragraph{Evaluation on the Node Classification Task}

In light of the success of GrePool+, we recognized the potential for combining the drop-with-uniform-loss strategy with the Graph Attention Network (GAT)~\cite{gat} to address the node classification task. To our knowledge, GAT aggregates information from all neighborhoods using attention mechanisms. However, not all the information is beneficial for node classification, especially for heterophilic graph datasets~\cite{h2gcn}.  To mitigate this issue, we applied the proposed drop-with-uniform-loss strategy to GAT (\textit{i.e.}, GAT+). This strategy effectively reduces the impact of noise information by discontinuing information aggregation from nodes with lower attention scores. Additionally, we applied uniform loss to the representations of these stop-aggregating nodes. To evaluate the effectiveness of GAT+, we conducted experiments on six commonly-used heterophilic datasets, including Cornell, Texas, Wisconsin, Actor, Squirrel, and Chameleon. The results presented in Table~\ref{tab:nc} consistently demonstrate that our method outperforms the baseline models across all datasets. This highlights the effectiveness and generalization ability of our proposed strategy and offers a novel perspective on node classification tasks.

\section{Conclusion}
This study introduced GrePool, an innovative graph pooling method designed to selectively discard nodes based on their direct impact on the final prediction outcome. This is achieved without additional networks or parameters. Based on this, we present GrePool+, an enhanced version of GrePool, which utilizes information from the nodes typically overlooked and discarded by standard graph pooling methods. This approach refines the training process and enhances classification accuracy. After that, we theoretically and empirically validated the efficacy and generalization capabilities of GrePool and GrePool+, providing substantial evidence for their applications. The experimental evaluation results demonstrate the effectiveness of our proposed methods and reveal several insightful observations regarding existing graph pooling practices. These findings hold the potential to inspire further advancements in the field. 

Despite the above contributions, this study encountered various challenges. Future research could focus on adjusting the attention weights, particularly concerning scores from different heads within the self-attention mechanism. Additionally, applying GrePool and GrePool+ to other graph-related tasks presents an opportunity for further exploration and validation of these methods.

\Acknowledgements{This work was supported in part by the Natural Science Foundation of China (Nos. 61976162, 82174230), Artificial Intelligence Innovation Project of Wuhan Science and Technology Bureau (No.2022010702040070).}



\bibliographystyle{plain}
\bibliography{reference}



\end{document}